# OrienText: Surface Oriented Textual Image Generation

Shubham Paliwal
TCS Research
New Delhi, INDIA
shubham.p3@tcs.com

Arushi Jain
TCS Research
New Delhi, INDIA
j.arushi@tcs.com

Monika Sharma
TCS Research
New Delhi, INDIA
monika.sharma1@tcs.com

Vikram Jamwal
TCS Research
Pune, INDIA
vikram.jamwal@tcs.com

Lovekesh Vig
TCS Research
New Delhi, INDIA
lovekesh.vig@tcs.com

## Abstract

Textual content in images is crucial in e-commerce sectors, particularly in marketing campaigns, product imaging, advertising, and the entertainment industry. Current text-to-image (T2I) generation diffusion models, though proficient at producing high-quality images, often struggle to incorporate text accurately onto complex surfaces with varied perspectives, such as angled views of architectural elements like buildings, banners, or walls. In this paper, we introduce the Surface *Orien*ted *Text*ual Image Generation **(OrienText)** method, which leverages region-specific surface normals as conditional input to T2I generation diffusion model. Our approach ensures accurate rendering and *correct orientation* of the text within the image context. We demonstrate the effectiveness of the OrienText method on a self-curated dataset of images and compare it against the existing textual image generation methods.

## 1 Introduction

Textual images are crucial for e-commerce, advertising, marketing, and entertainment industries. Creating *text content* into images requires specialized skills and advanced editing tools such as [Adobe Inc. 2023]. If integrated as a post-processing step for diffusion model-generated content, this approach may cause design inconsistencies and production scaling challenges. Therefore, GenAI-based methods for generating more precisely controlled image text are essential.

Recent years have witnessed remarkable progress in T2I generation using diffusion models [Betker and Goh 2023; Rombach et al. 2022; Ruiz et al. 2023; Song et al. 2023]. However, these methods often fall short of accurately generating textual images, leading to challenges such as spelling mistakes and lack of control over font attributes. Some new techniques promise to address these challenges [Chen et al. 2024, 2023; Ma et al. 2023; Paliwal et al. 2024; Tuo et al. 2023; Zhang et al. 2024]. Despite the advancements, they still fail to render text accurately on complex surfaces with angled views such as billboards, buildings, banners, and shipping containers. The resulting text appears overlaid (or frontal) and fails to blend harmoniously with the scene. For example, as in Fig. 1, when we attempt to replace the text "EAT" with "BUY" in the source image (S), previous methods generate the output image (a) where the text does not conform to the underlying surface orientation.

We propose that incorporating surface information can improve model control over text orientation. To achieve this, we use surface normals from the background as explicit conditional inputs for the T2I generation model. Recent advances have shown that convolutional networks can infer surface normals from a single image. We adopt a similar approach based on per-pixel ray direction, as outlined by [Bae and Davison 2024], where surface normals are learned by analyzing the rotation between neighboring pixels. This allows us to align individual characters of the character mask with the surface geometry, ensuring seamless text integration. The resulting output image, depicted in Fig. 1(b2), shows text that correctly aligns with the surface.

In summary, our work makes the following **contributions**:

(1) We introduce the method *OrienText*, which leverages surface normals of underlying geometry to generate correctly oriented text.
(2) We transform the character mask used in the specialized text-generating diffusion models to align each character's bounding box with the surface normals.
(3) We introduce a surface-normal consistency-based error metric to evaluate the alignment of text with the intended region.
(4) We identify harmonization, text rendering, and perspective blending as the key quality parameters and demonstrate performance of our approach on a diversified dataset of images of signboards, buildings, and product packaging.

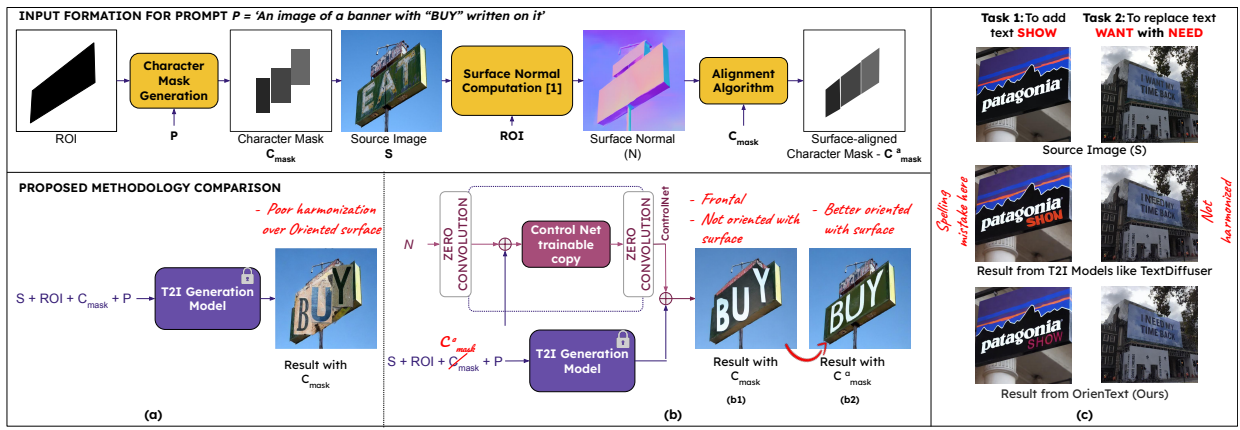

Figure 1: Proposed Workflow: (a) Traditional T2I methods such as TextDiffuser [Chen et al. 2023] use Source image (S), Text-prompt (P), Region-of-Interest (ROI), and Character Mask ($C_{mask}$) and generate an image where the text often appears overlaid and frontal, lacking proper alignment with the background surface. (b) Our proposed OrienText method takes Surface Normal (N) as an additional input to the ControlNet-augmented T2I method. The initial output image (b1) shows some improvement over (a), but the text is still not well-oriented with the surface. To address this, we transform $C_{mask}$ to align with the surface normals (N), resulting in $C_{mask}^a$ and producing a perfectly aligned and well-harmonized image in (b2). (c) Some example visual text generation results. Refer to Supplementary for more visual text generations.

## 2 Related Work

Diffusion models have emerged as powerful tools in generative modeling, particularly for image synthesis. Denoising Diffusion Probabilistic Models (DDPMs) introduced by Ho et al.[Ho et al. 2020] iteratively refine images through noise reduction, resulting in high-fidelity image generation. Models like DALLE-2[Ramesh et al. 2022], Imagen [Saharia et al. 2022], and Stable Diffusion [Rombach et al. 2022] leverage the semantic richness inherent in textual prompts to achieve impressive results.

*Textual Image Generation:* Integrating text within images while ensuring text-to-surface alignment accuracy and seamless blending remains a critical challenge. Early methods focused on overlaying text on images without considering the orientation of the underlying surface geometry. Yang et al. [Yang et al. 2016] introduced algorithms for blending text with complex backgrounds using texture and color analysis. Recent diffusion-based methods like GlyphDraw [Ma et al. 2023], TextDiffuser [Chen et al. 2024, 2023], and Brush-Your-Text [Zhang et al. 2024] have improved text-background harmonization. Generative models like Glyph-ByT5[Liu et al. 2024] use language models to encode text attributes but often struggle with perspective distortion, leading to misaligned and unnatural text integration on varied surfaces.

*Surface Normal Estimation:* Accurate surface normal estimation is crucial for computer vision tasks like 3D reconstruction and object recognition. Techniques by Eigen [Eigen and Fergus 2015] use CNNs to predict surface normals from an image, while recent work by Sun et al.[Sun et al. 2022] refines these predictions using multi-view stereo and depth sensing.

Combining surface normals with diffusion models offers a novel approach to integrating text into images. Our method uses diffusion models guided by surface normal estimations to align text with surface geometry, effectively mitigating perspective misalignment and producing visually coherent results.

## 3 Methodology: OrienText

Given an input source image (S) along with its corresponding text-generation region-of-interest (ROI) and character mask ($C_{mask}$), our aim is to generate the same image with the specified text perfectly oriented with the underlying surface, as shown in Fig. 1(b2). The text-generation ROI indicates the region in the image where the text is to be generated, while $C_{mask}$ marks the spatial positions of the individual characters with bounding boxes. Fig. 1(b) illustrates the workflow of our proposed method *OrienText*. It utilizes the ControlNet model which is designed to conditionally control the T2I generation models. To generate visually coherent textual images where the text is perfectly aligned with the underlying surface, we propose using the normal to the background surface as an input to the ControlNet model. This approach enables the base diffusion model such as TextDiffuser model [Chen et al. 2023] to generate images with text seamlessly integrated onto surfaces with angled views. Our method consists of the following components:

### 3.1 Finding Surface Normals

We first compute the surface normals of the given masked region, where the text needs to be appended, using the method proposed by Bae et. al. [Bae and Davison 2024]. This method utilizes the per-pixel ray direction and estimates the surface normal from a single image by learning the rotation between the neighboring pixels. Using the computed surface normals, we create a surface normal map where individual pixels are mapped with the normal vectors across the three axes, as shown by image N in Fig. 1.

### 3.2 Character Mask Alignment with Surface-Normals

To align character bounding boxes with surface normals, we define a projection plane perpendicular to the surface normal vector:

$$N = (n_x, n_y, n_z) \quad (1)$$

where N is the surface normal vector, and x, y, and z represent the three axes. Due to the absence of depth information, we assume a unit depth ($d = 1$) along the negative direction of the normal vector. We then compute the center of each bounding box ($C$) in 3D space, as defined in Eq. 2.

$$C = (c_x, c_y, c_z) \quad (2)$$

Next, we translate $C$ by the unit depth along the normal vector to project it onto the projection plane, resulting in a new center point $C'$ as defined in Eq. 3.

$$C' = (c_x - n_x, c_y - n_y, c_z - n_z) \quad (3)$$

Subsequently, we project $C'$ onto the plane defined by $n$ and the origin. To determine this projection, we find the intersection of the line passing through $C'$ and normal to the plane. The projected point $C_p$ is calculated using Eq. 4, where t is defined by Eq. 5.

$$C_p = C' + tN \quad (4)$$

$$t = \frac{n_x c_x + n_y c_y + n_z c_z}{n_x^2 + n_y^2 + n_z^2} \quad (5)$$

To preserve the width $w$ and height $h$ of the bounding box during projection, we determine the $2D$ coordinates of the bounding box corners relative to $C_p$. Thus, the four corners are computed by translating $C_p$ by $\pm \frac{w}{2}$ in the x-direction and $\pm \frac{h}{2}$ in the y-direction. Using the Eq. 1-5, we obtain the projected coordinates of the bounding box for textual characters, aligned with the projection surface, thereby ensuring accurate text placement.

### 3.3 ControlNet augmented T2I generation

Finally, we develop a ControlNet-augmented specialized T2I generation model that integrates surface normal as a control input with the pre-trained T2I model such as TextDiffuser. This model utilizes the source image ($S$), Region of interest ($ROI$) and character mask $C_{mask}$. The $ROI$ is a binary mask indicating the regions to be edited, while $C_{mask}$ encodes the character details in the pixel space at required spatial positions. Since image editing regions are often oriented at angles, we transform $C_{mask}$ to align with surface normals, resulting in $C_{mask}^a$. This transformation ensures a seamless $ROI$ surface, which cannot be achieved with the original square bounding boxes in $C_{mask}$, as shown in Fig. 1(b).

## 4 Experiments and Results
### 4.1 Experimental Setup

We create training dataset from the SCUT dataset [Lyu and Liao 2018] images and generated surface normals [Bae and Davison 2024]. Further, we apply affine transformations to augment the dataset, resulting in a total of 2,320 images. For training, we used the AdamW optimizer with a learning rate of $5e − 5$. The training was conducted with a batch size of 2, accumulating gradients every 4 batches over 60,000 training steps on a single NVIDIA V100 GPU. During inference, users specify the text region, and the model aligns the character mask with detected surface normals for image generation. For low-cost inference, we used FP16 precision, allowing the model to run efficiently on GPUs with less than 10GB of memory.

Table 1: Automated Quantitative Comparison Using Surface-Normal Consistency based Mean Angular Error (MAE-N)

| Method | w/o Mask Alignment MAE-N | w Mask Alignment MAE-N |
|---|---|---|
| TextDiffuser | 4.5243 | 3.9960 |
| TextDiffuser-2 | 2.2982 | 1.8832 |
| Anytext | 1.8937 | 1.8955 |
| **OrienText (Ours)** | 2.1486 | 1.7411 |

*Evaluation Dataset:* Since a suitable publicly available dataset of visual text images was not available for evaluation purposes, we curated a dataset of 60 images from the internet. These images include textual content over a variety of street signboards, building names, product packaging, and other relevant visual content. This dataset was used to evaluate our OrienText method.

### 4.2 Evaluation Approach:

We introduce a new metric based on surface-normal consistency for automatic evaluation and performed a human evaluation to assess the quality of generated images.

*4.2.1 Automated Evaluation:* Currently, no metric exists for automatically evaluating text orientation with background surfaces. To address this, we propose a surface-normal consistency metric. Effective text integration should maintain surface normal consistency before and after text generation. We first compute the surface normals [Bae and Davison 2024] of the image before and after text generation using our proposed OrienText method. This step is crucial as it allows us to assess any changes in the geometric representation of the surfaces caused by the text integration process. We then calculate the Mean Angular Error between the surface normals of the input and generated images, denoted as:
MAE-Normal = mean( cosine_similarity(N, N')), where $N \in \mathbb{R}^{h \times w \times 3}$ and $N' \in \mathbb{R}^{h \times w \times 3}$ represent the surface normals before and after the image generation, respectively. A lower MAE-N indicates that the surface normals remain similar, implying that the text characters are well-aligned with the underlying surface geometry. This quantitative analysis should validate the success of any approach in generating high-quality, perspective-aware textual images.

*4.2.2 Human Evaluation:* We conducted a survey for qualitative evaluation by human participants. The survey involved comparing image editing tasks performed by TextDiffuser [Chen et al. 2024], TextDiffuser-2 [Chen et al. 2023], AnyText [Tuo et al. 2023] and proposed OrienText method. The participants assessed the images based on the following three metrics:

- **Harmonization**: It evaluates how well the generated text regions blend harmoniously with the image regions.
- **Text Rendering Quality**: It assesses the quality of the rendered text, ensuring it is free from any noise or artifacts.
- **Perspective Blending**: This metric measures how accurately the generated text aligns with and conforms to the surface geometry of the region of interest (ROI).

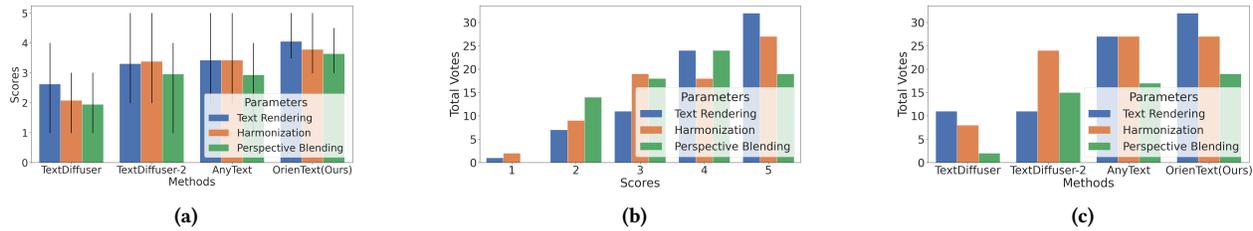

Figure 2: Human Evaluation Results: (a) Average scores and corresponding variances across three quality parameters. (b) Distribution ratings (1-5) for *OrienText* across 3 parameters (c) Distribution of the highest-rating(5) across 3 parameters.

The survey involved 15 participants with diverse backgrounds, who were unaware of the source algorithm for the generated images, which were randomized. Initially, participants viewed examples of both high and low-quality textual images and then rated each generated image on a scale of 1 to 5. To reduce evaluation discrepancies and ensure relative scores, we displayed four images from different methods on the same page (see Supplementary for a snapshot).

### 4.3 Results and Discussion

Table 1 presents the results of the surface-normal consistency metric. Our ablation study shows that aligning character masks with surface normals improves performance across all methods. Notably, our OrienText method consistently achieved the lowest MAE-N loss, demonstrating its superior ability to preserve surface normals after text generation. Next, we provide qualitative results from the human evaluations. Fig. 2a shows scores for each method across Text Rendering, Harmonization, and Perspective Blending, with higher scores indicating better performance. Our OrienText method outperforms others in all parameters, especially in Perspective Blending, where it achieves the highest score with low variance. This suggests that OrienText excels at integrating text into images and maintaining correct orientation. Fig. 2b shows the distribution OrienText ratings (1-5), with the highest votes for 5-rating for Text Rendering and Harmonization, and 4-rating for Perspective Blending, indicating potential for further improvement for the latter. Furthermore, Fig. 2c shows the distribution of 5-rating votes for different methods across three image quality parameters, highlighting that our method received the most 5-rating votes, indicating superior performance. Finally, qualitative visual examples are shown in Fig. 1(c). For more results and an example use-case of product advertisement image generation, (refer to Supplementary).

### 5 Conclusion

We introduced OrienText, a novel method for generating textual images that utilizes surface normals to determine text orientation, ensuring each character's bounding box aligns with the surface geometry. This approach addresses the shortcomings of current models that struggle with text alignment on complex angled surfaces. While OrienText automates text placement and enhances the visual coherence and realism of generated images, it may still encounter difficulties with very small fonts and 3D engraved texts. Nonetheless, our method significantly advances state-of-the-art techniques in text rendering, harmonization, and perspective blending. Future research will focus on overcoming these specific challenges related to small and engraved text rendering in images.

## References

Adobe Inc. 2023. *Adobe Photoshop*.

Gwangbin Bae and Andrew J Davison. 2024. Rethinking inductive biases for surface normal estimation. In *Proceedings of the IEEE/CVF CVPR*. 9535–9545.

James Betker and Gabriel et al. Goh. 2023. Improving image generation with better captions. *Computer Science* 2 (2023), 3.

Jingye Chen, Yupan Huang, Tengchao Lv, Lei Cui, and Qifeng Chen. 2024. Textdiffuser: Diffusion models as text painters. *Advances in NeurIPS* 36 (2024).

Jingye Chen, Yupan Huang, Tengchao Lv, Lei Cui, Qifeng Chen, and Furu Wei. 2023. TextDiffuser: Diffusion Models as Text Painters. *arXiv* (2023).

David Eigen and Rob Fergus. 2015. Predicting depth, surface normals and semantic labels with a common multi-scale convolutional architecture. In *Proceedings of the IEEE international conference on computer vision*. 2650–2658.

Jonathan Ho, Ajay Jain, and Pieter Abbeel. 2020. Denoising diffusion probabilistic models. *Advances in neural information processing systems* 33 (2020), 6840–6851.

Zeyu Liu, Weicong Liang, Zhanhao Liang, Chong Luo, Ji Li, Gao Huang, and Yuhui Yuan. 2024. Glyph-ByT5: A Customized Text Encoder for Accurate Visual Text Rendering. *arXiv preprint arXiv:2403.09622* (2024).

Pengyuan Lyu and Minghui et al. Liao. 2018. Mask textspotter: An end-to-end trainable neural network for spotting text with arbitrary shapes. In *ECCV*. 67–83.

Jian Ma, Mingjun Zhao, Chen Chen, and Ruichen Wang. 2023. GlyphDraw: Learning to Draw Chinese Characters in Image Synthesis Models Coherently. *arXiv preprint* (2023).

Shubham Paliwal, Arushi Jain, Monika Sharma, Vikram Jamwal, and Lovekesh Vig. 2024. CustomText: Customized Textual Image Generation using Diffusion Models. *arXiv preprint arXiv:2405.12531* (2024).

Aditya Ramesh, Prafulla Dhariwal, Alex Nichol, Casey Chu, and Mark Chen. 2022. Hierarchical text-conditional image generation with clip latents. *arXiv preprint arXiv:2204.06125* 1, 2 (2022), 3.

Robin Rombach, Andreas Blattmann, Dominik Lorenz, Patrick Esser, and Björn Ommer. 2022. High-resolution image synthesis with latent diffusion models. In *IEEE/CVF conference on computer vision and pattern recognition*. 10684–10695.

Nataniel Ruiz, Yuanzhen Li, Varun Jampani, Yael Pritch, Michael Rubinstein, and Kfir Aberman. 2023. Dreambooth: Fine tuning text-to-image diffusion models for subject-driven generation. In *IEEE/CVF CVPR*. 22500–22510.

Chitwan Saharia, William Chan, Saurabh Saxena, Lala Li, Jay Whang, Emily L Denton, Kamyar Ghasemipour, Raphael Gontijo Lopes, Burcu Karagol Ayan, Tim Salimans, et al. 2022. Photorealistic text-to-image diffusion models with deep language understanding. *Advances in NeurIPS* 35 (2022), 36479–36494.

Yang Song, Prafulla Dhariwal, Mark Chen, and Ilya Sutskever. 2023. Consistency models. *arXiv preprint arXiv:2303.01469* (2023).

Shang Sun, Dan Xu, Hao Wu, Haocong Ying, and Yurui Mou. 2022. Multi-view stereo for large-scale scene reconstruction with MRF-based depth inference. *Computers & Graphics* 106 (2022), 248–258.

Yuxiang Tuo, Wangmeng Xiang, Jun-Yan He, Yifeng Geng, and Xuansong Xie. 2023. Anytext: Multilingual visual text generation and editing. *arXiv preprint arXiv:2311.03054* (2023).

Xuyong Yang, Tao Mei, Ying-Qing Xu, Yong Rui, and Shipeng Li. 2016. Automatic generation of visual-textual presentation layout. *ACM Transactions on Multimedia Computing, Communications, and Applications (TOMM)* 12, 2 (2016), 1–22.

Lingjun Zhang, Xinyuan Chen, and Yaohui Wang. 2024. Brush your text: Synthesize any scene text on images via diffusion model. In *AAAI*, Vol. 38. 7215–7223.

# Supplementary: OrienText: Surface Oriented Textual Image Generation

Shubham Paliwal, Arushi Jain, Monika Sharma, Vikram Jamwal, Lovekesh Vig
{shubham.p3,j.arushi,monika.sharma1,vikram.jamwal,lovekesh.vig}@tcs.com
TCS Research, India

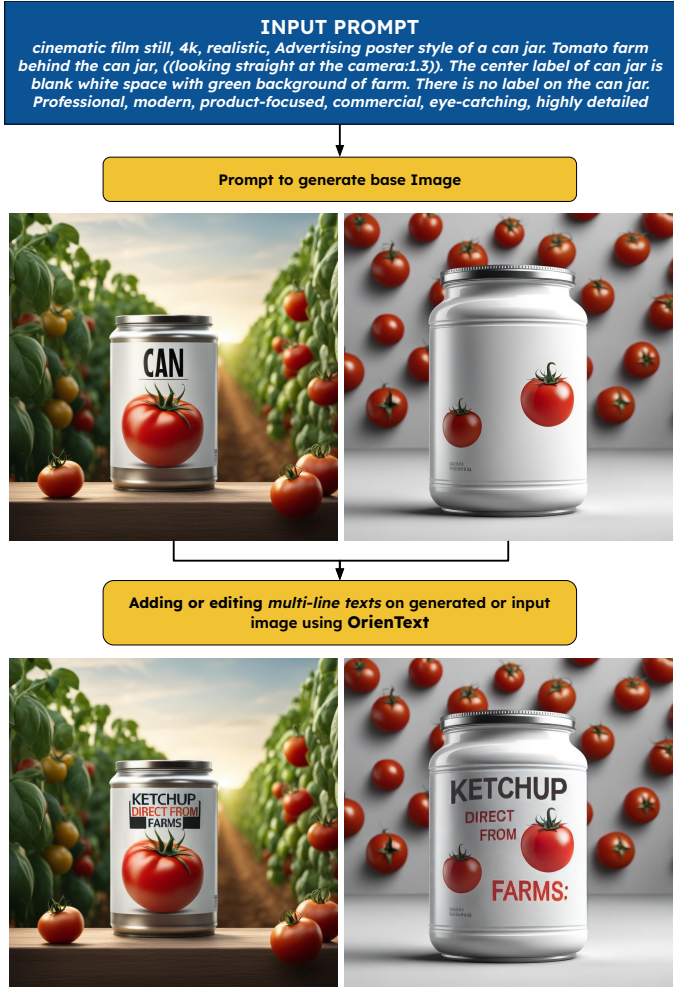

Figure 1: Example workflow for generating a product advertisement image: We first generate a base image using a diffusion model, then customize it with OrienText by adding or editing the required multi-line text labels.

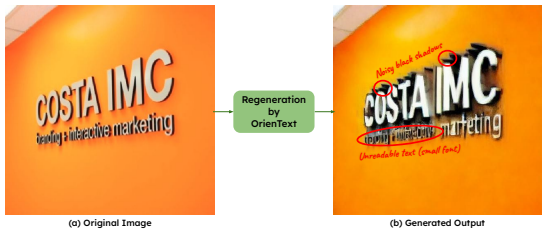

Figure 2: An example image highlighting OrienText's limitations in generating 3D text and small fonts. The rendered 3D text shows noisy shadows, and the small-sized text is unreadable.

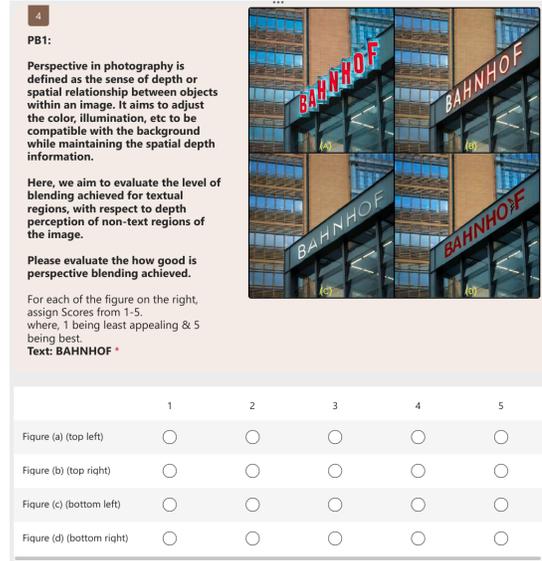

Figure 3: Screenshot of the survey interface where participants evaluated images based on three parameters. The text 'BAHNHOF' was generated using four methods, each producing a distinct image, displayed side-by-side in random order to ensure unbiased scoring.

# References

[1] Jingye Chen, Yupan Huang, Tengchao Lv, Lei Cui, and Qifeng Chen. 2024. Textdiffuser: Diffusion models as text painters. *Advances in NeurIPS* 36 (2024).
[2] Jingye Chen, Yupan Huang, Tengchao Lv, Lei Cui, Qifeng Chen, and Furu Wei. 2023. TextDiffuser: Diffusion Models as Text Painters. *arXiv* (2023).
[3] Yuxiang Tuo, Wangmeng Xiang, Jun-Yan He, Yifeng Geng, and Xuansong Xie. 2023. Anytext: Multilingual visual text generation and editing. *arXiv preprint arXiv:2311.03054* (2023).

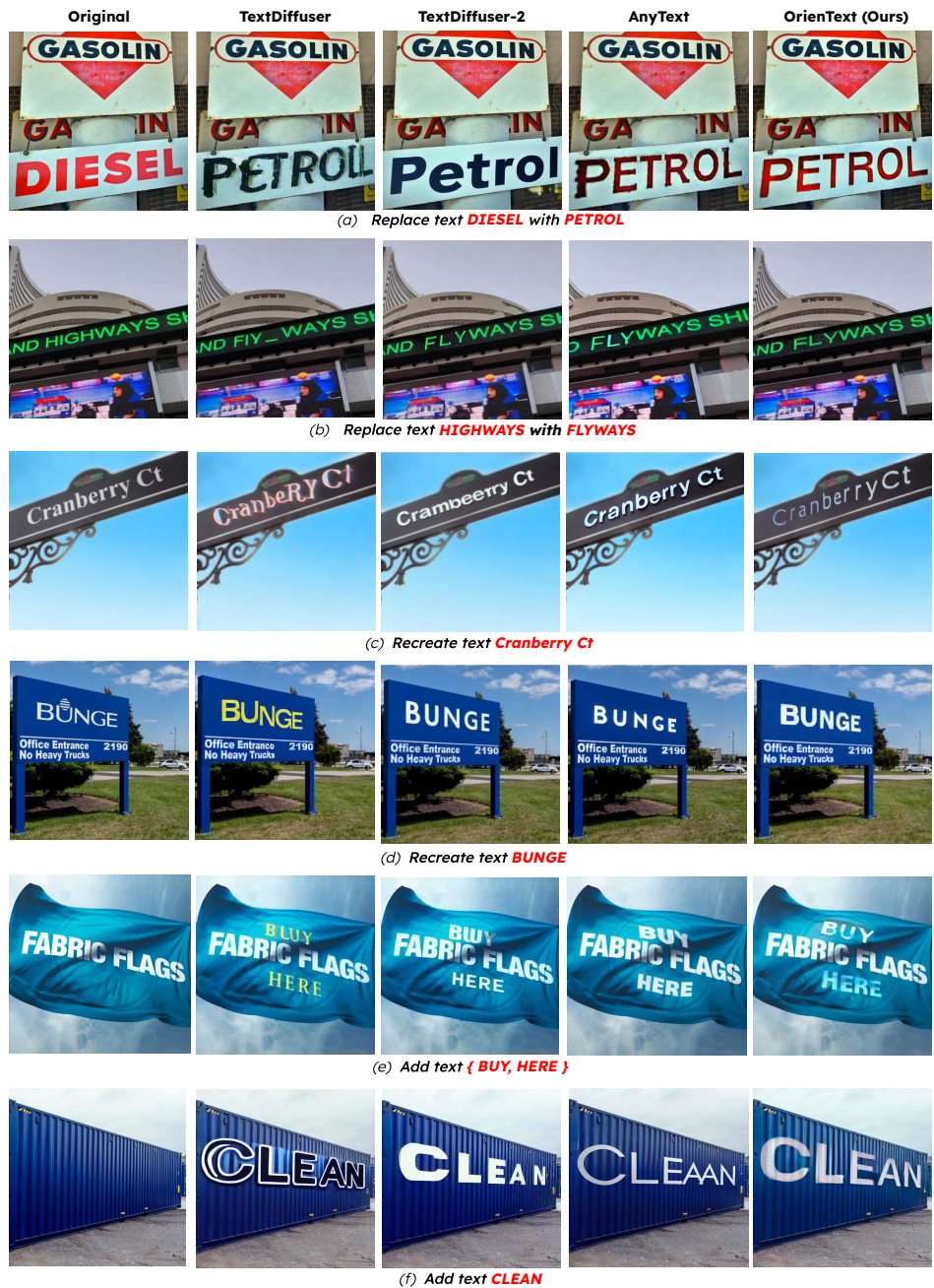

Figure 4: Qualitative comparison of generated images using different methods such as TextDiffuser [2], TextDiffuser-2 [1], AnyText [3] and our proposed approach OrienText for disparate image editing tasks.